\documentclass[11pt,a4paper]{article}
\usepackage[hyperref]{emnlp-ijcnlp-2019}
\usepackage{times}
\usepackage{latexsym}
\usepackage{multirow}
\usepackage{amssymb, amsmath}
\usepackage{url}
\usepackage{graphicx}
\usepackage{booktabs}
\usepackage{chngcntr}

\aclfinalcopy 

\newcommand{\fragment}{FLC }
\newcommand{\sentence}{SLC }

\newcommand{\fragmentend}{FLC}
\newcommand{\sentenceend}{SLC}

\newcommand{\del}[1]{}

\newcommand{\Ni}{({\em i})~}
\newcommand{\Nii}{({\em ii})~}

\title{Fine-Grained Analysis of Propaganda in News Articles}

\author{\textrm{Giovanni Da San Martino}$^{1}$ \quad \textrm{Seunghak Yu}$^{2}$ \quad Alberto Barr\'{o}n-Cede\~{n}o$^{3}$ \\ \textbf{ Rostislav Petrov}$^{4}$ \quad \textbf{Preslav Nakov}$^{1}$ \\ $^{1}$ Qatar Computing Research Institute, HBKU, Qatar \\   
$^{2}$ MIT Computer Science and Artificial Intelligence Laboratory,
Cambridge, MA, USA  \\ 
$^{3}$ Universit\`{a} di Bologna, Forl\`{i}, Italy, \textrm{}$^{4}$ A Data Pro, Sofia, Bulgaria \\
{\tt \{gmartino, pnakov\}@hbku.edu.qa} 
  \\ {\tt seunghak@csail.mit.edu}, {\tt a.barron@unibo.it}
  \\ {\tt rostislav.petrov@adata.pro}}
 
\date{}
 
\begin{document}
\maketitle
\begin{abstract}

Propaganda aims at influencing people's mindset with the purpose of advancing a specific agenda. Previous work has addressed propaganda detection at the document level, typically labelling \emph{all} articles from a propagandistic news outlet as propaganda. Such noisy gold labels inevitably affect the quality of any learning system trained on them. A further issue with most existing systems is the lack of explainability. 
To overcome these limitations, we propose a novel task: performing fine-grained analysis of texts by detecting all fragments that contain propaganda techniques as well as their type. In particular, we create a corpus of news articles manually annotated at the fragment level with eighteen propaganda techniques and we propose a suitable evaluation measure.  
We further design a novel multi-granularity neural network, and we show that it outperforms several strong BERT-based baselines. 
\end{abstract}

\section{Introduction} \label{sec:intro}

Research on detecting propaganda has focused primarily on articles~\cite{BARRONCEDENO20191849,rashkin-EtAl:2017:EMNLP2017}.
In many cases, there are no labeled data for individual articles, but there are such labels for entire news outlets. Thus, often all articles from the same news outlet get labeled the way that this outlet is labeled. 
Yet, it has been observed that propagandistic sources could post objective non-propagandistic articles periodically to increase their credibility~\cite{Horne2018}. Similarly, media generally recognized as objective might occasionally post articles that promote a particular editorial agenda and are thus propagandistic.
Thus, it is clear that transferring the label of the news outlet to each of its articles, could introduce noise. 
Such labels can still be useful for training robust systems, but they cannot be used to get a fair assessment of a system at testing time. 

\noindent One option to deal with the lack of labels for articles is to crowdsource the annotation. 
However, in preliminary experiments we observed that the average annotator cannot detach her personal mindset from the judgment of propaganda and bias, i.e.,~if a clearly propagandistic text expresses ideas aligned with the annotator's beliefs, it is unlikely that she would judge it as such.

We argue that in order to study propaganda in a sound and reliable way, we need to rely on high-quality trusted professional annotations and 
it is best to do so at the fragment level, targeting specific techniques rather than using a label for an entire document or an entire news outlet. 

Ours is the first work that goes at a fine-grained level: identifying specific instances of propaganda techniques used within an article. 
In particular, we create a corresponding corpus. 
For this purpose, we asked six experts to annotate articles from news outlets recognized as propagandistic and non-propagandistic, marking specific text spans with eighteen propaganda techniques. We also designed appropriate evaluation measures.
Taken together, the annotated corpus and the evaluation measures represent the first manually-curated evaluation framework for the analysis of fine-grained propaganda. 
We release the corpus (350K tokens) as well as our code in order to enable future research.\footnote{The corpus, the evaluation measures, and the models are available at \url{http://propaganda.qcri.org/}}
Our contributions are as follows:

\begin{itemize}  \setlength\itemsep{0pt}
    \item We formulate a new problem: detect the use of specific propaganda techniques in text.
    \item We build a new large corpus for this problem.
    \item We propose a suitable evaluation measure.
    \item We design a novel multi-granularity neural network, and we show that it outperforms several strong BERT-based baselines.
\end{itemize}

\noindent Our corpus could enable research in propagandistic and non-objective news, including the development of explainable AI systems. A system that can detect instances of use of specific propagandistic techniques would be able to make it explicit to the users why a given article was predicted to be propagandistic. It could also help train the users to spot the use of such techniques in the news.

The remainder of this paper is organized as follows: Section~\ref{sec:background} presents the propagandistic techniques we focus on. Section~\ref{sec:data} describes our corpus. Section~\ref{sec:evaluation} discusses an evaluation measures for comparing labeled fragments.  Section~\ref{sec:models} presents the formulation of the task and our proposed models. Section~\ref{sec:experiments} describes our experiments and the evaluation results. Section~\ref{sec:relatedwork} presents some relevant related work. Finally, Section~\ref{sec:conclusions} concludes and discusses future work.

\section{Propaganda and its Techniques}
\label{sec:background}

Propaganda comes in many forms, but it can be recognized by its persuasive function, sizable target audience, the representation of a specific group's agenda, and the use of faulty reasoning and/or emotional appeals~\cite{Miller}. 
Since propaganda is conveyed through the use of a number of techniques, their detection allows for a deeper analysis at the paragraph and the sentence level that goes beyond a single document-level judgment on whether a text is propagandistic.

Whereas the definition of propaganda is widely accepted in the literature, 
the set of propaganda techniques differs between scholars~\cite{Torok2015}. For instance, 
\citet{Miller} considers seven techniques, whereas~\citet{Weston2000} lists at least 24, and Wikipedia discusses 69.\footnote{\url{http://en.wikipedia.org/wiki/Propaganda_techniques}; last visit May 2019.} 
The differences are mainly due to some authors ignoring some techniques, or using definitions that subsume the definition used by other authors.
Below, we describe the propaganda techniques we consider: a curated list of eighteen items derived from the aforementioned studies. The list only includes techniques that can be found in journalistic articles and can be judged intrinsically, without the need to retrieve supporting information from external resources. For example, we do not include techniques such as \emph{card stacking}~\cite[page 237]{Jowett2012a}, since it would require comparing against external sources of information. 

\noindent The eighteen techniques we consider are as follows (cf.\ Table~\ref{tab:instances} for examples): 

\begin{table*}[tbh]
\small
\centering
\begin{tabular}{l @{\hspace{1mm}}   p{0.87\textwidth}}
\toprule
\multicolumn{1}{c}{\bf Doc ID} & \bf Technique	$\bullet$ Snippet \\
\midrule
783702663 & \texttt{loaded language}	$\bullet$ until forced to act by \textbf{a worldwide storm of outrage}.			\\
732708002 & \texttt{name calling, labeling}	$\bullet$ dismissing the protesters as “\textbf{lefties}” and hugging Barros publicly	\\
701225819 & \texttt{repetition}			$\bullet$ 	Farrakhan repeatedly refers to Jews as “\textbf{Satan}.” He states to his audience [\ldots] call them by their real name, `\textbf{Satan}.'\\
782086447 & \texttt{exaggeration, minimization}	$\bullet$ heal the situation of \textbf{extremely grave} immoral behavior	\\
761969038 & \texttt{doubt}			$\bullet$ \textbf{Can the same be said for the Obama Administration}?	\\
696694316 & \texttt{appeal to fear/prejudice}	$\bullet$ \textbf{A dark, impenetrable and “irreversible” winter of persecution of the faithful by their own shepherds will fall}.	\\
776368676 & \texttt{flag-waving}			$\bullet$ conflicted, and \textbf{his 17 Angry Democrats that are doing his dirty work are a disgrace to USA}! ---Donald J. Trump	\\
776368676 & \texttt{flag-waving}			$\bullet$ attempt (Mueller) \textbf{to stop the will of We the People}!!! It's time to jail Mueller 	\\
735815173 & \texttt{causal oversimplification}	$\bullet$ he said \textbf{The people who talk about the "Jewish question" are generally anti-Semites}. Somehow I don't think	\\
781768042 & \texttt{causal oversimplification}	$\bullet$ will not be reversed, \textbf{which leaves no alternative as to why God judges and is judging America today }	\\
111111113 & \texttt{slogans}			$\bullet$ \textbf{BUILD THE WALL!"} Trump tweeted.\\
783702663 & \texttt{appeal to authority}		$\bullet$ \textbf{Monsignor Jean-François Lantheaume, who served as first Counsellor of the Nunciature in Washington, confirmed that ``Viganò said the truth. That’s all''}	\\
783702663 & \texttt{black-and-white fallacy}	$\bullet$ Francis said these words: “\textbf{Everyone is guilty for the good he could have done and did not do \ldots If we do not oppose evil, we tacitly feed it}. 	\\
729410793 & \texttt{thought-terminating cliches}	$\bullet$ \textbf{I do not really see any problems there.} Marx is the President	\\
770156173 & \texttt{whataboutism}		$\bullet$ President Trump ---\textbf{who himself avoided national military service} in the 1960's--- keeps beating the war drums over North Korea	\\ 
778139122 & \texttt{reductio ad hitlerum}	$\bullet$ ``Vichy journalism,'' a term which now fits so much of the mainstream media. \textbf{It collaborates in the same way that the Vichy government in France collaborated with the Nazis.}	\\
778139122 & \texttt{red herring}			$\bullet$ It describes the tsunami of vindictive personal abuse that has been heaped upon Julian from well-known journalists, many claiming liberal credentials. The Guardian, \textbf{which used to consider itself the most enlightened newspaper in the country}, has probably been the worst.	\\
698018235 & \texttt{bandwagon}			$\bullet$ He tweeted, ``\textbf{EU no longer considers \#Hamas a terrorist group. Time for US to do same.''} 	\\
729410793 & obfusc., int. vagueness, confusion $\bullet$ \textbf{The cardinal's office maintains that rather than saying ``yes,'' there is a possibility of liturgical ``blessing'' of gay unions, he answered the question in a more subtle way without giving an explicit ``yes.''}	\\
783702663 & \texttt{straw man}			$\bullet$ ``Take it seriously, but with a large grain of salt.'' \textbf{Which is just Allen's more nuanced way of saying: ``Don't believe it}.''	\\
\bottomrule
\end{tabular}
\caption{Instances of the different propaganda techniques from our corpus. We show the document ID, the technique, and the text snippet, in bold. When necessary, some context is provided to better understand the example. \label{tab:instances}}
\end{table*}

\paragraph{1. Loaded language.}
Using words/phrases with strong emotional implications (positive or negative) to influence an audience~\cite[p.~6]{Weston2000}.
\textit{Ex.:} ``[\ldots] a lone lawmaker's childish shouting.''

\paragraph{2. Name calling or labeling.}
Labeling the object of the propaganda campaign as either something the target audience fears, hates, finds undesirable or otherwise loves or praises~\cite{Miller}. \textit{Ex.:} ``Republican congressweasels'', ``Bush the Lesser.''

\paragraph{3. Repetition.} 
Repeating the same message over and over again, so that the audience will eventually accept it~\cite{Torok2015,Miller}.

\paragraph{4. Exaggeration or minimization.}
Either representing something in an excessive manner: making things larger, better, worse (e.g., ``the best of the best'', ``quality guaranteed'') 
or making something seem less important or smaller than it actually is~\cite[p.~303]{Jowett2012a}, e.g., saying that an insult was just a joke.
\textit{Ex.:} ``Democrats bolted as soon as Trump’s speech ended in an apparent effort to signal they can't even stomach being in the same room as the president''; ``I was not fighting with her; we were just playing.''

\paragraph{5. Doubt.}
Questioning the credibility of someone or something. \textit{Ex.:} A candidate says about his opponent: ``Is he ready to be the Mayor?''

\paragraph{6. Appeal to fear/prejudice.} 
Seeking to build support for an idea by instilling anxiety and/or panic in the population towards an alternative, possibly based on preconceived judgments.
\textit{Ex.:} ``stop those refugees; they are terrorists.''

\paragraph{7. Flag-waving.} 
Playing on strong national feeling (or with respect to a group, e.g.,~race, gender, political preference) to justify or promote an action or idea \cite{Hobbs2008}.
\textit{Ex.:} ``entering this war will make us have a better future in our country.''

\paragraph{8. Causal oversimplification.} Assuming one cause when there are multiple causes behind an issue. 
We include \emph{scapegoating} as well: the transfer of the blame to one person or group of people without investigating the complexities of an issue. \textit{Ex.:} ``If France had not declared war on Germany, World War II would have never happened.''

\paragraph{9. Slogans.}
A brief and striking phrase that may include labeling and stereotyping. Slogans tend to act as emotional appeals~\cite{As2015}.
\textit{Ex.:} ``Make America great again!''

\smallskip
\noindent \textbf{10. Appeal to authority.}
Stating that a claim is true simply because a valid authority/expert on the issue supports it, without any other supporting evidence~\cite{Goodwin2011}. We include the special case where the reference is not an authority/expert, although it is referred to as \emph{testimonial} in the literature~\cite[p.~237]{Jowett2012a}.

\smallskip
\noindent \textbf{11. Black-and-white fallacy, dictatorship.}
Presenting two alternative options as the only possibilities, when in fact more possibilities exist \cite{Torok2015}. As an extreme case, telling the audience exactly what actions to take, eliminating any other possible choice (\emph{dictatorship}).
\textit{Ex.:} ``You must be a Republican or Democrat; you are not a Democrat. Therefore, you must be a Republican''; ``There is no alternative to war.''

\smallskip
\noindent \textbf{12. Thought-terminating \textit{clich\'e}.}
Words or phrases that discourage critical thought and meaningful discussion about a given topic. They are typically short, generic sentences that offer seemingly simple answers to complex questions or that distract attention away from other lines of thought~\cite[p.~78]{Hunter2015}.
\textit{Ex.:} ``it is what it is''; ``you cannot judge it without experiencing it''; ``it's common sense'', ``nothing is permanent except change'', ``better late than never''; 
``mind your own business''; ``nobody's perfect''; ``it doesn't matter''; ``you can't change human nature.''

\smallskip
\noindent \textbf{13. Whataboutism.} Discredit an opponent's position by charging them with hypocrisy without directly disproving their argument~\cite{Richter2017}. For example, mentioning an event that discredits the opponent: ``What about \ldots?''~\cite{Richter2017}.
\textit{Ex.:} Russia Today had a proclivity for whataboutism in its coverage of the 2015 Baltimore and Ferguson protests in the US, which revealed a consistent refrain: ``the oppression of blacks in the US has become so unbearable that the eruption of violence was inevitable'', and that the US therefore lacks ``the moral high ground to discuss human rights issues in countries like Russia and China.''

\smallskip
\noindent \textbf{14. Reductio ad Hitlerum.} 
Persuading an audience to disapprove an action or idea by suggesting that the idea is popular with groups hated in contempt by the target audience. It can refer to any person or concept with a negative connotation~\cite{Aper2009}.
\textit{Ex.:}  ``Only one kind of person can think this way: a communist.''

\smallskip
\noindent \textbf{15. Red herring.}
Introducing irrelevant material to the issue being discussed, so that everyone's attention is diverted away from the points made~\cite[p.\ 78]{Weston2000}. 
Those subjected to a red herring argument are led away from the issue that had been the focus of the discussion and urged to follow an observation or claim that may be associated with the original claim, but is not highly relevant to the issue in dispute~\cite{Aper2009}.
Ex.: ``You may claim that the death penalty is an ineffective deterrent against crime -- but what about the victims of crime? How do you think surviving family members feel when they see the man who murdered their son kept in prison at their expense? Is it right that they should pay for their son's murderer to be fed and housed?''

\smallskip
\noindent \textbf{16. Bandwagon.}
Attempting to persuade the target audience to join in and take the course of action because ``everyone else is taking the same action''~\cite{Hobbs2008}.
Ex.: ``Would you vote for Clinton as president? 57\% say yes.''

\smallskip
\noindent \textbf{17. Obfuscation, intentional vagueness, confusion.} 
Using deliberately unclear words, so that the audience may have its own interpretation~\cite[p.~8]{Suprabandari2007,Weston2000}. 
For instance, when an unclear phrase with multiple possible meanings is used within the argument, and, therefore, it does not really support the conclusion. Ex.: ``It is a good idea to listen to victims of theft. Therefore, if the victims say to have the thief shot, then you should do it.''

\smallskip
\noindent \textbf{18. Straw man.} When an opponent's proposition is substituted with a similar one which is then refuted in place of the original~\cite{Walton1996}. 
\citet[p.~78]{Weston2000} specifies the characteristics of the substituted proposition: ``caricaturing an opposing view so that it is easy to refute.''

We provided the above definitions, together with some examples and an annotation schema, to our professional annotators, so that they can manually annotate news articles. The details are provided in the next section.

\section{Data Creation} \label{sec:data}

We retrieved 451 news articles from 48 news outlets, both propagandistic and non-propagandistic, which we annotated as described below.

\subsection{Article Retrieval}\label{sec:articlecollection}

First, we selected 13 propagandistic and 36 non-propagandistic news media outlets, as labeled by Media Bias/Fact Check.\footnote{\url{http://mediabiasfactcheck.com/}} Then, we retrieved articles from these sources, as shown in Table~\ref{tab:statscorpus}.
Note that 82.5\% of the articles are from  propagandistic sources, and these articles tend to be longer.

Table~\ref{tab:propsources} shows the number of articles retrieved from each propagandistic outlet. Overall, we have 350k word tokens,
which is comparable to standard datasets for other fine-grained text analysis tasks, such as named entity recognition, e.g.,~CoNLL'02 and CoNLL'03 covered 381K, 333K, 310K, and 301K tokens
for Spanish, Dutch, German, and English, respectively  \cite{TjongKimSang:2002:ICS:1118853.1118877,TjongKimSang:2003:ICS:1119176.1119195}.

\begin{table}
\small
\centering
\begin{tabular}{p{2.9cm}rrr}
\toprule
		  & \bf Prop & \bf Non-prop & \bf All\\
\midrule
articles  & 372 & 79 & 451 \\
avg length (lines) & 49.8 & 34.4 & 47.1 \\
avg length (words) & 973.2 & 635.4 & 914.0 \\
avg length (chars) & 5,942 & 3,916 & 5,587 \\
\bottomrule
\end{tabular}
\caption{Statistics about the articles retrieved with respect to the category of the media source: \textbf{prop}agandistic, \textbf{non-prop}agandistic, and all together. \label{tab:statscorpus}}
\end{table}

\begin{table}
\small
\centering
\begin{tabular}{@{ }l@{ }r @{\hspace{2mm}} l@{ }r@{ }}
\toprule
\bf News Outlet & \bf \# 			& \bf News Outlet & \bf \# \\
\midrule
Freedom	Outpost		&	133	& The Remnant	Magazine &	14	\\
Frontpage Magazine	&	56	& Breaking911		&	11\\
shtfplan.com		&	55	& truthuncensored.net	&	8\\
Lew	Rockwell	&	26	& The Washington Standard	&	6\\
vdare.com		&	20	& www.unz.com		&	5\\
remnantnewspaper.com	&	19	& www.clashdaily.com	&	1\\
Personal Liberty	&	18	& \\

\bottomrule
\end{tabular}

\caption{Number of articles retrieved from news outlets deemed propagandistic by Media Bias/Fact Check. \label{tab:propsources}}

\end{table}

\subsection{Manual Annotation} \label{sec:annotationschemes}

\begin{figure}
	\includegraphics[width=0.49\textwidth]{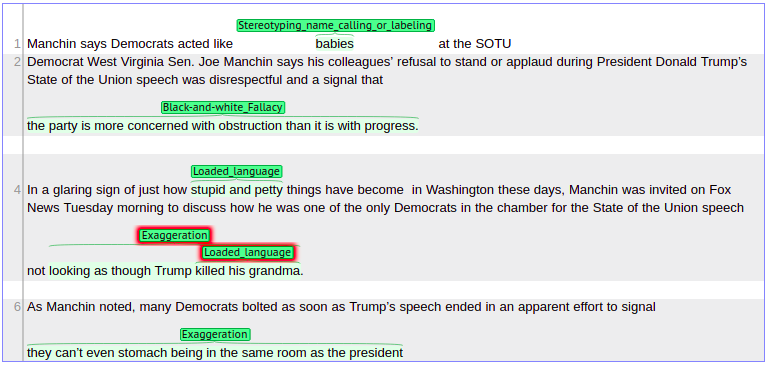}
    \caption{Example given to the annotators. \label{fig:example}}
\vspace{-6pt}
\end{figure}

We aim at obtaining text fragments annotated with any of the 18 techniques described in Section~\ref{sec:background} (see Figure~\ref{fig:example} for an example). 
Since the time required to understand and memorize all the propaganda techniques is significant, this annotation task is not well-suited for crowdsourcing. 
We partnered instead with a company that performs professional annotations, A Data Pro.\footnote{\url{http://www.aiidatapro.com}}
Appendix A shows details about the instructions and the tools provided to the annotators.

We computed the $\gamma$ inter-annotator agreement~\cite{Mathet2015}.
We chose $\gamma$ because
\Ni it is designed for tasks where both the span and its label are to be found and
\Nii it can deal with overlaps in the annotations by the same annotator\footnote{See~\cite{meyer-EtAl:2014:ColingDemo,Mathet2015} for other alternatives, which lack some properties; \Nii in particular.}
(e.g., instances of \textit{doubt} often use \textit{name calling} or \textit{loaded language} to reinforce their message).
We computed $\gamma_{s}$, where we only consider the identified spans, regardless of the technique, and $\gamma_{sl}$, where we consider both the spans and their labels. 

Let $a$ be an annotator. In a preliminary exercise, four annotators $a_{[1,..,4]}$ annotated six articles independently, and the agreement was $\gamma_{s}=0.34$ and $\gamma_{sl}=0.31$. Even taking into account that $\gamma$ is a pessimistic measure~\cite{Mathet2015}, these values are low. Thus, we designed an annotation schema composed of two stages and involving two annotator teams, each of which covered about $220$ documents. In stage 1, both $a_1$ and $a_2$ annotated the same documents independently. In stage 2, they gathered with a consolidator $c_1$ to discuss all instances and to come up with a final annotation. Annotators $a_3$ and $a_4$ and consolidator $c_2$ followed the same procedure.
Annotating the full corpus took 395 man hours. 

\begin{table}[t]
\small
\centering
\begin{tabular}{cc cc}
\toprule
\multicolumn{2}{l}{\bf Annotations} & \bf spans ($\gamma_s$)	& \bf +labels ($\gamma_{sl}$)\\
\midrule
$a_1$	& $a_2$	& 0.30	& 0.24	\\
$a_3$	& $a_4$	& 0.34	& 0.28	\\\midrule
$a_1$	& $c_1$	& 0.58	& 0.54	\\
$a_2$	& $c_1$	& 0.74	& 0.72	\\
$a_3$	& $c_2$	& 0.76	& 0.74	\\
$a_4$	& $c_2$	& 0.42	& 0.39	\\
\bottomrule
\\
\end{tabular}
\caption{$\gamma$ inter-annotator agreement between annotators spotting spans alone (\textbf{spans}) and spotting spans+labeling (\textbf{+labels}). The top-2 rows refer to the first stage: agreement between annotators. The bottom 4 rows refer to the consolidation stage: agreement between each annotator and the final gold annotation. \label{tab:agreement}}
\vspace{-6pt}
\end{table}

Table~\ref{tab:agreement} shows the $\gamma$ agreements on the full corpus. As in the preliminary annotation, the agreements for both teams are relatively low: 0.30 and 0.34 for span selection, and slightly lower when labeling is considered as well. 
After the annotators discussed with the consolidator on the disagreed cases, the $\gamma$ values got much higher: up to 0.74 and 0.76 for each team. 
We further analyzed the annotations to determine the main cause for the disagreement by computing the percentage of instances spotted by one annotator only in the first stage that are retained as gold annotations. 

Overall the percentage is $53\%$ ($5,921$ out of $11,122$), and for each annotator is  $a_1=70\%$, $a_2=48\%$, $a_3=57\%$, $a_4=31\%$. 
Observing such percentages together with the relatively low differences in Table~\ref{tab:agreement} between $\gamma_s$ and $\gamma_{sl}$ for the same pairs $(a_i, a_j)$ and $(a_i, c_j)$, we can conclude that disagreements are in general not due to the two annotators assigning different labels to the same or mostly overlapping spans, but rather because one has missed an instance in the first stage. 

\subsection{Statistics about the Dataset} \label{sec:corpusstats}

The total number of technique instances found in the articles, after the consolidation phase, is $7,485$, with respect to a total number of $21,230$ sentences (35.2\%). 
Table~\ref{tab:techniquestats} reports some statistics about the annotations. 
The average propagandistic fragment has a length of $47$ characters and the average length of a sentence is 112.5 characters. 

On average, the propagandistic techniques are half a sentence long. 
The most common ones are \textit{loaded language} and \textit{name calling, labeling} with $2,547$ and $1,294$ occurrences, respectively. 
They appear 6.7 and 4.7 times per article, while no other technique appears more than twice. 
Note that repetition are inflated as we asked the annotators to mark both the original and the repeated instances. 

\begin{table}[t]
\small
\centering
\begin{tabular}{p{4.3cm}@{\hspace{-1mm}}r@{\hspace{2mm}} r}
\toprule
\bf Propaganda Technique	& \bf inst & \bf avg. length\\
\midrule
loaded language			& 2,547	& $23.70\pm 25.30$	\\
name calling, labeling		& 1,294	& $26.10\pm 19.88$	\\
repetition			& 767	& $16.90\pm 18.92$	\\
exaggeration, minimization	& 571	& $45.36\pm 35.55$	\\
doubt				& 562	& $123.21\pm 97.65$	\\
appeal to fear/prejudice	& 367	& $93.56\pm 74.59$	\\
flag-waving			& 330	& $61.88\pm 68.61$	\\
causal oversimplification	& 233	& $121.03\pm 71.66$	\\
slogans				& 172	& $25.30\pm 13.49$\\
appeal to authority		& 169	& $131.23\pm 123.2$	\\
black-and-white fallacy		& 134	& $98.42\pm 73.66$	\\
thought-terminating cliches	& 95	& $34.85\pm 29.28$	\\
whataboutism			& 76	& $120.93\pm 69.62$	\\ 
reductio ad hitlerum		& 66	& $94.58\pm 64.16$	\\
red herring			& 48	& $63.79\pm 61.63$	\\
bandwagon			& 17	& $100.29\pm 97.05$	\\
obfusc., int. vagueness, confusion & 17	& $107.88\pm 86.74$	\\
straw man			& 15	& $79.13\pm 50.72$	\\
\midrule
\bf all				& 7,485	& $46.99\pm 61.45$	\\
\bottomrule
\end{tabular}
\caption{Corpus statistics including \textbf{inst}ances per technique and their \textbf{avg. length} in terms of characters. \label{tab:techniquestats}}
\end{table}

\section{Evaluation Measures}
\label{sec:evaluation}

Our task is a sequence labeling one, with the following key characteristics: (\textit{i})~a large number of techniques whose spans might overlap in the text, and (\textit{ii})~large lengths of these spans.
This requires an evaluation measure that gives credit for partial overlaps.\footnote{The evaluation measures for the CoNLL'02 and CoNLL'03 NER tasks, where an instance is considered properly identified if and only if both the boundaries and the label are correct~\cite{Tsai:06}, are not suitable in our context.}
We derive an \textit{ad hoc} measure following related work on named entity recognition (NER)~\cite{Nadeau:07} and (intrinsic) plagiarism detection (PD)~\cite{Potthast2010a}. 

While in NER, the relevant fragments tend to be short multi-word strings, in PD ---and in our propaganda technique identification task--- the length varies widely (cf.\ Table~\ref{tab:techniquestats}), and instances span from single tokens to full sentences or even longer pieces of text. 
Thus, in our precision and recall versions,
we give partial credit to imperfect matches at the character level, as in PD.

\begin{figure}[tbh]
	\includegraphics[width=\columnwidth]{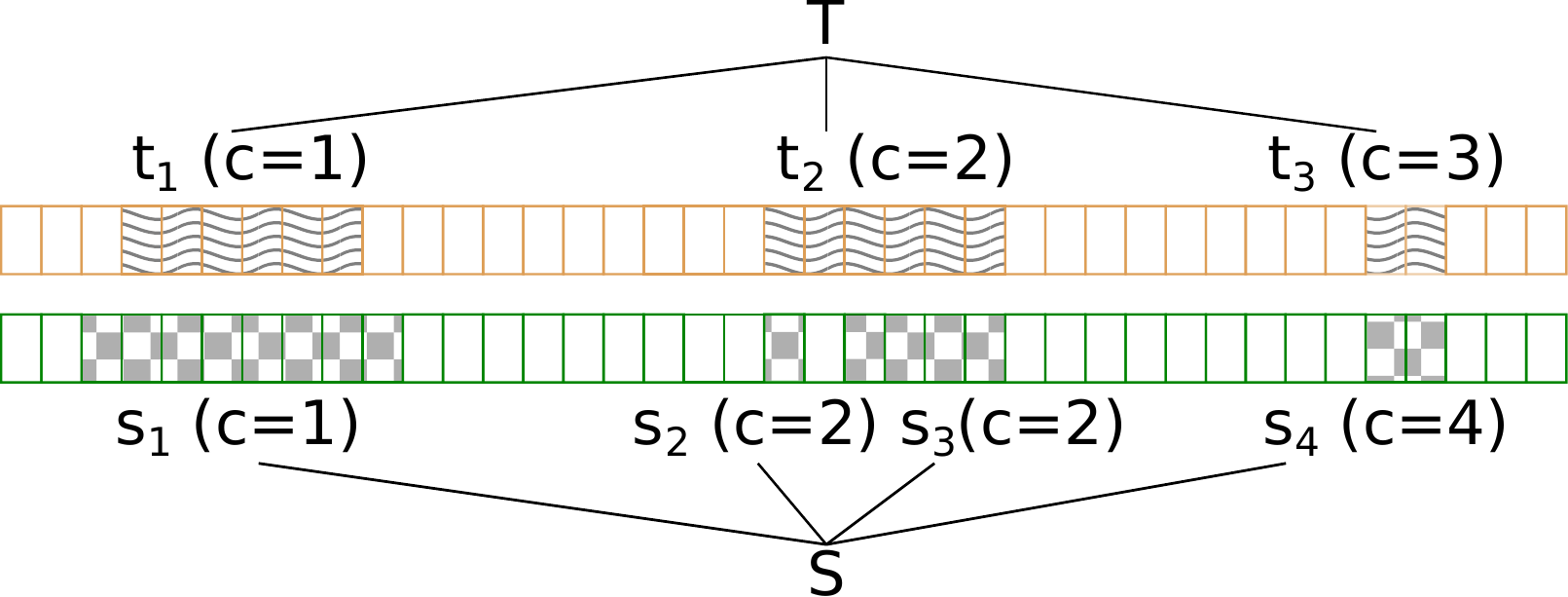}
    \caption{Example of gold annotation (top) and the predictions of a supervised model (bottom) in a document represented as a sequence of characters. The class of each fragment is shown in parentheses. $s_1$ goes beyond $t_1$'s proper boundaries; $s_2$ and $s_3$ partially spot $t_2$, but fail to identify it entirely; $s_4$ spots the exact boundaries of $t_3$, but fails to assign it the right label. \label{fig:metrics_example}}
\end{figure}

Let document $\boldsymbol{d}$ be represented as a sequence of characters. 
A propagandistic text fragment is then represented as $t=[t_i, \ldots, t_j]\subseteq \boldsymbol{d}$. 
A document includes a set of (possibly overlapping) fragments $T$. 
Similarly, a learning algorithm produces a set $S$ with fragments $s=[s_m, \dots, s_n]$, predicted on $\boldsymbol{d}$. 
A labeling function $l(x) \in \{1,\ldots,18\}$ associates $s\in S$ to one of the eighteen techniques. 
Figure~\ref{fig:metrics_example} gives examples of gold and predicted fragments. 

We define the following function to handle partial overlaps between fragments with same labels:
\begin{equation}
    C(s,t,h) = \frac{|(s\cap t)|}{h}\delta\left(l(s), l(t) \right), 
    \label{eq:annotationsScore}
\end{equation}
where $h$ is a normalizing factor and $\delta(a,b)=1$ if $a=b$, and $0$ otherwise. 
In the future, $\delta$ could be refined to account for custom distance functions between classes, e.g., we might consider mistaking \textit{loaded language} for \textit{name calling or labeling} less problematic than confusing it with \textit{Reduction ad Hitlerum}.  
Given Eq.~\eqref{eq:annotationsScore}, we now define variants of precision and recall able to account for the imbalance in the corpus:
\begin{equation}
P(S,T) = \frac{1}{|S|}\!\sum_{\begin{minipage}[t]{0.97cm}\footnotesize $s\in S,\\$ $t\in T$\end{minipage}}\!\!\! C(s,t,|s|), 
\label{eq:precisiontask3}
\end{equation}
\begin{equation}
R(S,T) = \frac{1}{|T|}\!\sum_{\begin{minipage}[t]{0.97cm}\footnotesize $s\in S,\\$ $t\in T$\end{minipage}}\!\!\! C(s,t,|t|),
\label{eq:recalltask3}
\end{equation}

We define Eq.~\eqref{eq:precisiontask3} to be zero if $|S|=0$ and Eq.~\eqref{eq:recalltask3} to be zero if $|T|=0$. 
Following~\citet{Potthast2010a}, in Eqs.~\eqref{eq:precisiontask3} and~\eqref{eq:recalltask3} we penalize systems predicting too many or too few instances by dividing by $|S|$ and $|T|$, respectively, e.g., in Figure~\ref{fig:metrics_example} $P(\{s_2, s_3\}, T) < P(\{s_3\}, T)$. 
Finally, we combine Eqs.~\eqref{eq:precisiontask3} and \eqref{eq:recalltask3} into an F$_1$-measure, the harmonic mean of precision and recall.

Having a separate function $C$ to be responsible for comparing two annotations gives us some additional flexibility that is missing in standard NER measures that operate at the token/character level. 
For example, in Eq.~\eqref{eq:annotationsScore} we could easily change the factor that gives credit for partial overlaps by being more forgiving when only few characters are wrong.

\section{Tasks and Proposed Models}
\label{sec:models}

We define two tasks based on the corpus described in Section~\ref{sec:data}: \Ni \textbf{\sentence (Sentence-level Classification)}, which asks to predict whether a sentence contains at least one propaganda technique, and \Nii \textbf{\fragment (Fragment-level classification)}, which asks to identify both the spans and the type of propaganda technique. 
Note that these two tasks are of different granularities, $g_1$ and $g_2$, i.e.,~tokens for \fragment and sentences for \sentenceend. 
We split the corpus into training, development and test, each containing 293, 57, 101 articles and 14,857, 2,108, 4,265 sentences.

\subsection{Baselines} 
\label{sec:baselines}

We depart from BERT~\cite{devlin2018bert}, as it has achieved state-of-the-art performance on multiple NLP benchmarks, and we design three baselines based on it.

\textbf{BERT.} We add a linear layer on top of BERT and we fine-tune it, as suggested in~\cite{devlin2018bert}. For the \fragment task, we feed the final hidden representation for each token to a layer $L_{g_2}$ that makes a 19-way classification: does this token belong to one of the eighteen propaganda techniques or to none of them (cf.\ Figure~\ref{fig:arch}-a).
For the \sentence task, we feed the final hidden representation for the special \verb![CLS]! token, which BERT uses to represent the full sentence, to a two-dimensional layer $L_{g_1}$ to make a binary classification.

\textbf{BERT-Joint.} We use the layers for both tasks in the BERT baseline, $L_{g_1}$ and $L_{g_2}$, and we train for both \fragment and \sentence jointly (cf.\ Figure~\ref{fig:arch}-b).

\textbf{BERT-Granularity.} 
We modify BERT-Joint to transfer information from \sentence directly to \fragmentend. Instead of using only the $L_{g_2}$ layer for \fragmentend, we concatenate $L_{g_1}$ and $L_{g_2}$, and we add an extra 19-dimensional classification layer $L_{g_{1,2}}$ on top of that concatenation to perform the prediction for \fragment (cf.\ Figure~\ref{fig:arch}-c).

\subsection{Multi-Granularity Network}
\label{sub:mgn}

\begin{figure}[tbh]
\centering
\includegraphics[width=\columnwidth]{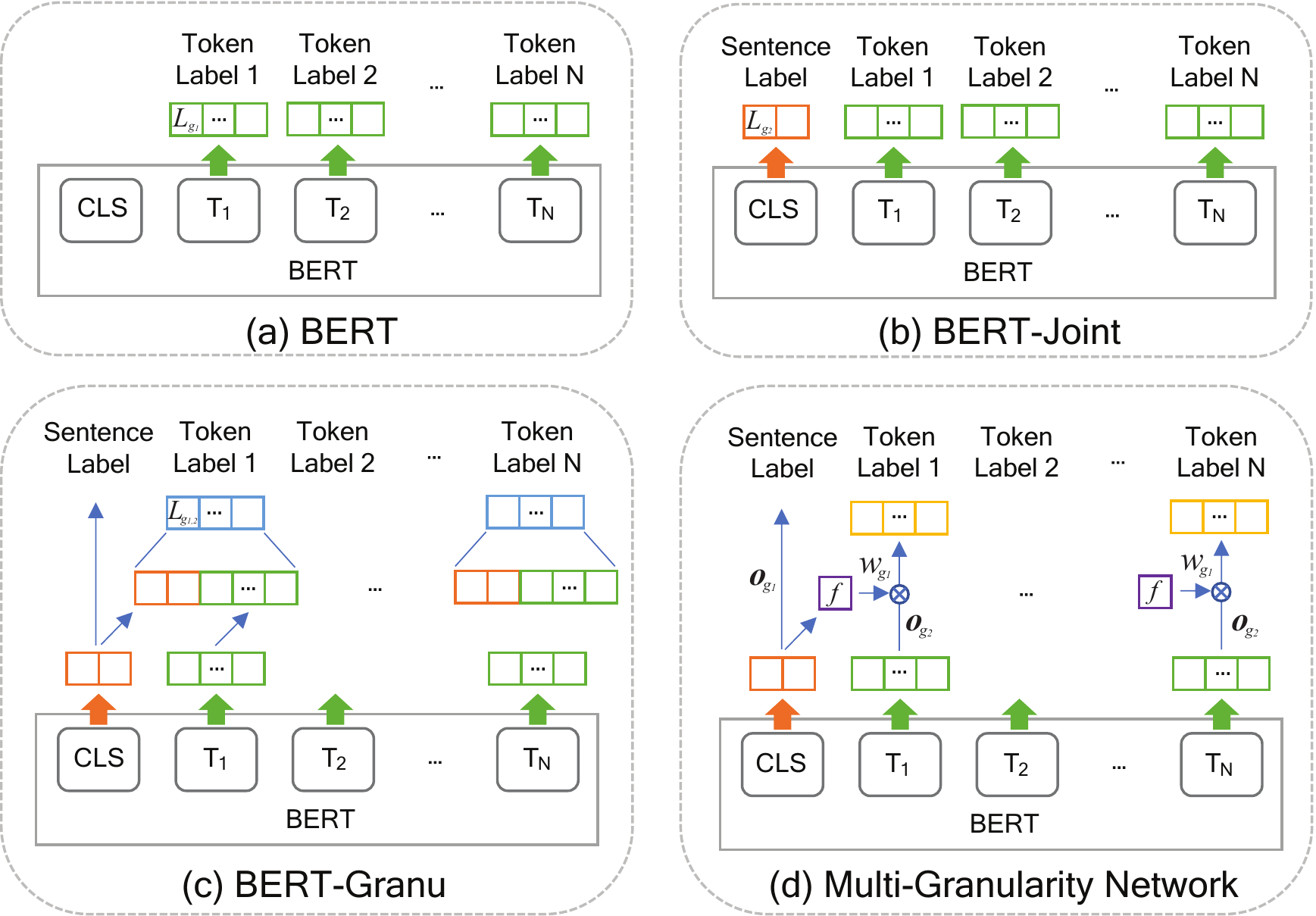}
\caption{The architecture of the baseline models (a-c), and of our proposed multi-granularity network (d).}
\label{fig:arch}
\end{figure}

We propose a model that can drive the higher-granularity task (\fragmentend) on the basis of the lower-granularity information (\sentenceend), rather than simply using low-granularity information directly. Figure~\ref{fig:arch}-d shows the architecture of this model.
More generally, suppose there are $k$ tasks of increasing granularity, e.g., document-level, paragraph-level, sentence-level, word-level, subword-level, character-level.

\noindent Each task has a separated classification layer $L_{g_k}$ that receives the feature representation of the specific level of granularity $g_k$ and outputs $\boldsymbol{o}_{g_k}$.
The dimension of the representation depends on the embedding layer, while the dimension of the output depends on the number of classes in the task. The output $\boldsymbol{o}_{g_k}$ generates a weight for the next granularity task $g_{k+1}$ through a trainable gate $f$:

\begin{equation}
w_{g_{k}} = f(\boldsymbol{o}_{g_k})
\end{equation}

The gate $f$ consists of a projection layer to one dimension and an activation function. The resulting weight is multiplied by each element of the output of layer $L_{g_{k+1}}$ to produce the output for task $g_{k+1}$:

\begin{equation}
\boldsymbol{o}_{g_{k+1}} = w_{g_{k}} * \boldsymbol{o}_{g_{k+1}}
\end{equation}

If $w_{g_{k}}=0$ for a given example, the output of the next granularity task $\boldsymbol{o}_{g_{k+1}}$ would be 0 as well. In our setting this means that, if the sentence-level classifier is confident the sentence does not contain propaganda, i.e.,~$w_{g_{k}}=0$, then $\boldsymbol{o}_{g_{k+1}}=0$ and there would be no propagandistic technique predicted for any span within that sentence. 
Similarly, when back-propagating the error, 
if $w_{g_{k}}=0$ for a given example, the final entropy loss would become zero; i.e.\ the model would not get any information from that example. As a result, only examples strongly classified as negative in a lower-granularity task would be ignored in the high-granularity task. 
Having the lower-granularity as the main task means that higher-granularity information can be selectively used as additional information to improve the performance, but only if the example is not considered as highly negative. We show this in Section~\ref{sub:spd}.

\noindent For the loss function, we use a cross-entropy loss with sigmoid activation for every layer, except for the highest-granularity layer $L_{g_K}$, which uses a cross-entropy loss with softmax activation. Unlike softmax, which normalizes over all dimensions, the sigmoid allows each output component of layer $L_{g_k}$ to be independent from the rest. Thus, the output of the sigmoid for the positive class increases the degree of freedom by not affecting the negative class, and vice versa. As we have two tasks, we use sigmoid activation for $L_{g_1}$ and softmax activation for $L_{g_2}$. Moreover, we use a weighted sum of losses with a hyper-parameter $\alpha$:

\begin{equation}
\mathcal{L}_\mathcal{J} = \mathcal{L}_{g_1} * \alpha + \mathcal{L}_{g_2} * (1-\alpha) 
\end{equation}

Again, we use BERT~\cite{devlin2018bert} for the contextualized embedding layer and we place the multi-granularity network on top of it.

\section{Experiments and Evaluation}
\label{sec:experiments}

\subsection{Experimental Setup}

We used the PyTorch framework and the pretrained BERT model, which we fine-tuned for our tasks.
We trained all models  using the following hyper-parameters: batch size of 16, sequence length of 210, weight decay of 0.01, and early stopping on validation F$_1$ with patience of 7. For optimization, we used Adam with a learning rate of 3e-5 and a warmup proportion of 0.1. To deal with class imbalance, we give weight to the binary cross-entropy according to the proportion of positive samples. For the $\alpha$ in the joint loss function, we use 0.9 for sentence classification, and 0.1 for word-level classification.
In order to reduce the effect of random fluctuations for BERT, all the reported numbers are the average of three experimental runs with different random seeds.
As it is standard, we tune our models on the dev partition and we report results on the test partition.

\subsection{Fragment-Level Propaganda Detection}

Table~\ref{Table:task3} shows the performance for the three baselines and for our multi-granularity network on the \fragment task. For the latter, we vary the degree to which the gate function is applied: using ReLU is more aggressive compared to using the Sigmoid, as the ReLU outputs zero for a negative input. Note that, even though we train the model to predict both the spans and the labels, we also evaluated it with respect to the spans only.

\begin{table}[t]
\centering
\begin{tabular}{@{}l@{\hspace{1mm}} c @{\hspace{1mm}}c@{\hspace{1mm}}c c@{\hspace{1mm}}c@{\hspace{1mm}}c@{}}
\toprule
 \multirow{2}{*}{\bf Model} & \multicolumn{3}{c}{\bf Spans} & \multicolumn{3}{c}{\bf Full Task} \\ 
 & \bf P & \bf R & \bf F$_1$ & \bf P & \bf R & \bf F$_1$ \\
 \midrule
BERT    &  39.57 & 36.42 & 37.90 & 21.48  & \bf 21.39 & 21.39\\
\,\,\, Joint      &  39.26 & 35.48 & 37.25 & 20.11 & 19.74 & 19.92\\
\,\,\, Granu      &  43.08 & 33.98 & 37.93   & 23.85 & 20.14 & 21.80\\
\multicolumn{2}{l}{\hspace{-3mm}Multi-Granularity} \\

\,\,\, ReLU    &  43.29 & 34.74  & 38.28   & 23.98 & 20.33 & 21.82\\
\,\,\, Sigmoid & \bf 44.12 & \bf 35.01 & \bf 38.98 & \bf 24.42 & 21.05 & \bf 22.58\\
\bottomrule
\end{tabular}
\caption{Fragment-level experiments (\fragment task). Shown are two evaluations: (\emph{i})~\textbf{Spans} checks only whether the model has identified the fragment spans correctly, while (\emph{ii})~\textbf{Full task} is evaluation wrt the actual task of identifying the spans and also assigning the correct propaganda technique for each span.
}
\label{Table:task3}
\end{table}

Table~\ref{Table:task3} shows that joint learning (BERT-Joint) hurts the performance compared to single-task BERT. 
However, using additional information from the sentence-level for the token-level classification (BERT-Granularity) yields small improvements. 
The multi-granularity models outperform all baselines thanks to their higher precision. This shows the effect of the model excluding sentences that it determined to be non-propagandistic from being considered for token-level classification. Nevertheless, the performance of sentence-level classification is far from perfect, achieving an F$_1$ of up to 60.98 (cf.\ Table~\ref{Table:task2}). The information it contributes to the final classification is noisy and the more conservative removal of instances performed by the Sigmoid function yields better performance than the more aggressive ReLU.

\subsection{Sentence-Level Propaganda Detection}
\label{sub:spd}

Table~\ref{Table:task2} shows the results for the \sentence task. We apply our multi-granularity network model to the sentence-level classification task to see its effect on low granularity when we train the model with a high granularity task. Interestingly, it yields huge performance improvements on the sentence-level classification result. Compared to the BERT baseline, it increases the recall by 8.42\%, resulting in a 3.24\% increase of the F$_1$ score. In this case, the result of token-level classification is used as additional information for the sentence-level task, and it helps to find more positive samples. This shows the opposite effect of our model compared to the \fragment task. Note also that using ReLU is more effective than using the Sigmoid, unlike in token-level classification. 

\begin{table}[tbh]
\centering
\begin{tabular}{l|ccc}
\toprule
\bf Model  & \bf Precision & \bf Recall & \bf F1 \\ 
 \midrule
 All-Propaganda  & 23.92 & 100.0 & 38.61\\
 BERT & \bf{63.20} & 53.16 & 57.74 \\ 
 BERT-Granu & 62.80 & 55.24 & 58.76 \\
 BERT-Joint & 62.84 & 55.46 & 58.91 \\
 MGN Sigmoid & 62.27 & 59.56 & 60.71 \\
 MGN ReLU & 60.41 & \textbf{61.58} & \textbf{60.98} \\
\toprule
\end{tabular}
\caption{Sentence-level (\sentenceend) results. \textit{All-propaganda} is a baseline which always output the propaganda class.}
\label{Table:task2}
\end{table}

\noindent Thus, since the performance range of the token-level classification is low, we think it is more effective to get additional information after aggressively removing negative samples by using ReLU as a gate in the model.

\section{Related Work}
\label{sec:relatedwork}

Propaganda identification has been tackled mostly at the article level.
\citet{rashkin-EtAl:2017:EMNLP2017} created a corpus of news articles labelled as belonging to four categories: propaganda, trusted, hoax, or satire. They included articles from eight sources, two of which are propagandistic. 
\citet{AAAI2019:proppy} experimented with a binarized version of the corpus from \cite{rashkin-EtAl:2017:EMNLP2017}: propaganda vs. the other three categories.
The corpus labels were obtained with distant supervision, assuming that all articles from a given news outlet share the label of that outlet, which inevitably introduces noise~\cite{Horne2018}. 

A related field is that of computational argumentation which, among others, deals with some logical fallacies related to propaganda. 
\citet{Habernal2018} presented a corpus of Web forum discussions with cases of \textit{ad hominem} fallacy identified. 
\citet{Habernal.et.al.2017.EMNLP,Habernal2018b} introduced \textit{Argotario}, a game to educate people to recognize and create fallacies. A byproduct of Argotario is a corpus with $1.3k$ arguments annotated with five fallacies, including \textit{ad hominem}, \textit{red herring} and \textit{irrelevant authority}, which directly relate to propaganda techniques (cf.\ Section~\ref{sec:background}).
Differently from~\cite{Habernal.et.al.2017.EMNLP,Habernal2018b,Habernal2018}, our corpus has 18 techniques annotated on the same set of news articles. Moreover, our annotations aim at identifying the minimal fragments related to a technique instead of flagging entire arguments.   

\section{Conclusion and Future Work} \label{sec:conclusions}

We have argued for a new way to study propaganda in news media: by focusing on identifying the instances of use of specific propaganda techniques. Going at this fine-grained level can yield more reliable systems and it also makes it possible to explain to the user why an article was judged as propagandistic by an automatic system. 

In particular, we designed an annotation schema of 18 propaganda techniques, and we annotated a sizable dataset of documents with instances of these techniques in use. We further designed an evaluation measure specifically tailored for this task. 
We made the schema and the dataset publicly available, thus facilitating further research. 
We hope that the corpus would raise interest outside of the community of researchers studying propaganda: the techniques related to fallacies and the ones relying on emotions might provide a novel setting for the researchers interested in Argumentation and Sentiment Analysis.

We experimented with a number of BERT-based models and devised a novel architecture which outperforms standard BERT-based baselines. 
Our fine-grained task can complement document-level judgments, both to come out with an aggregated decision and to explain why a  document ---or an entire news outlet--- has been flagged as potentially propagandistic by an automatic system. 

We are collaborating with A Data Pro to expand the corpus. In the mid-term, we plan to build an online platform where professors in relevant fields (e.g., journalism, mass communication) can train their students to recognize and annotate propaganda techniques. The hope is to be able to accumulate annotations as a by-product of using the platform for training purposes.

\section*{Acknowledgments}

This research is part of the Tanbih project,\footnote{\url{http://tanbih.qcri.org/}} which aims to limit the effect of ``fake news'', propaganda and media bias by making users aware of what they are reading. The project is developed in collaboration between the MIT Computer Science and Artificial Intelligence Laboratory (CSAIL) and the Qatar Computing Research Institute (QCRI), HBKU.

\bibliography{emnlp-ijcnlp-2019,propaganda,other}

\begin{thebibliography}{27}
\expandafter\ifx\csname natexlab\endcsname\relax\def\natexlab#1{#1}\fi

\bibitem[{Barr{\'o}n-Cede{\~n}o et~al.(2019)Barr{\'o}n-Cede{\~n}o,
  Da~San~Martino, Jaradat, and Nakov}]{AAAI2019:proppy}
Alberto Barr{\'o}n-Cede{\~n}o, Giovanni Da~San~Martino, Israa Jaradat, and
  Preslav Nakov. 2019.
\newblock Proppy: A system to unmask propaganda in online news.
\newblock In \emph{Proceedings of the 33rd AAAI Conference on Artificial
  Intelligence}, AAAI~'19, pages 9847--9848, Honolulu, HI, USA.

\bibitem[{Barr{\'o}n-Cedeno et~al.(2019)Barr{\'o}n-Cedeno, Jaradat,
  Da~San~Martino, and Nakov}]{BARRONCEDENO20191849}
Alberto Barr{\'o}n-Cedeno, Israa Jaradat, Giovanni Da~San~Martino, and Preslav
  Nakov. 2019.
\newblock Proppy: Organizing the news based on their propagandistic content.
\newblock \emph{Information Processing \& Management}, 56(5):1849--1864.

\bibitem[{Dan(2015)}]{As2015}
Lavinia Dan. 2015.
\newblock {Techniques for the Translation of Advertising Slogans}.
\newblock In \emph{{Proceedings of the International Conference Literature,
  Discourse and Multicultural Dialogue}}, LDMD~'15, pages 13--23, Mures,
  Romania.

\bibitem[{Devlin et~al.(2019)Devlin, Chang, Lee, and
  Toutanova}]{devlin2018bert}
Jacob Devlin, Ming-Wei Chang, Kenton Lee, and Kristina Toutanova. 2019.
\newblock {BERT}: Pre-training of deep bidirectional transformers for language
  understanding.
\newblock In \emph{Proceedings of the 2019 Conference of the North American
  Chapter of the Association for Computational Linguistics: Human Language
  Technologies}, NAACL-HLT~'19, pages 4171--4186, Minneapolis, MN, USA.

\bibitem[{Goodwin(2011)}]{Goodwin2011}
Jean Goodwin. 2011.
\newblock {Accounting for the force of the appeal to authority}.
\newblock In \emph{Proceedings of the 9th International Conference of the
  Ontario Society for the Study of Argumentation}, OSSA~'11, pages 1--9,
  Ontario, Canada.

\bibitem[{Habernal et~al.(2017)Habernal, Hannemann, Pollak, Klamm, Pauli, and
  Gurevych}]{Habernal.et.al.2017.EMNLP}
Ivan Habernal, Raffael Hannemann, Christian Pollak, Christopher Klamm, Patrick
  Pauli, and Iryna Gurevych. 2017.
\newblock Argotario: Computational argumentation meets serious games.
\newblock In \emph{Proceedings of the Conference on Empirical Methods in
  Natural Language Processing}, EMNLP~'17, pages 7--12, Copenhagen, Denmark.

\bibitem[{Habernal et~al.(2018{\natexlab{a}})Habernal, Pauli, and
  Gurevych}]{Habernal2018b}
Ivan Habernal, Patrick Pauli, and Iryna Gurevych. 2018{\natexlab{a}}.
\newblock Adapting serious game for fallacious argumentation to {G}erman:
  pitfalls, insights, and best practices.
\newblock In \emph{Proceedings of the Eleventh International Conference on
  Language Resources and Evaluation}, LREC~'18, Miyazaki, Japan.

\bibitem[{Habernal et~al.(2018{\natexlab{b}})Habernal, Wachsmuth, Gurevych, and
  Stein}]{Habernal2018}
Ivan Habernal, Henning Wachsmuth, Iryna Gurevych, and Benno Stein.
  2018{\natexlab{b}}.
\newblock Before name-calling: Dynamics and triggers of ad hominem fallacies in
  web argumentation.
\newblock In \emph{Proceedings of the 2018 Conference of the North American
  Chapter of the Association for Computational Linguistics: Human Language
  Technologies}, NAACL-HLT~'18, pages 386--396, New Orleans, LA, USA.

\bibitem[{Hobbs and Mcgee(2008)}]{Hobbs2008}
Renee Hobbs and Sandra Mcgee. 2008.
\newblock Teaching about propaganda: An examination of the historical roots of
  media literacy.
\newblock \emph{Journal of Media Literacy Education}, 6(62):56--67.

\bibitem[{Horne et~al.(2018)Horne, Khedr, and Adali}]{Horne2018}
Benjamin~D Horne, Sara Khedr, and Sibel Adali. 2018.
\newblock Sampling the news producers: A large news and feature data set for
  the study of the complex media landscape.
\newblock In \emph{International AAAI Conference on Web and Social Media},
  ICWSM~'18, Stanford, CA, USA.

\bibitem[{Hunter(2015)}]{Hunter2015}
John Hunter. 2015.
\newblock Brainwashing in a large group awareness training? {T}he classical
  conditioning hypothesis of brainwashing.
\newblock Master's thesis, University of Kwazulu-Natal, Pietermaritzburg, South
  Africa.

\bibitem[{Jowett and O'Donnell(2012)}]{Jowett2012a}
Garth~S. Jowett and Victoria O'Donnell. 2012.
\newblock What is propaganda, and how does it differ from persuasion?
\newblock In \emph{Propaganda \& Persuasion}, chapter~1, pages 1--48. Sage
  Publishing.

\bibitem[{Mathet et~al.(2015)Mathet, Widl{\"o}cher, and
  M{\'e}tivier}]{Mathet2015}
Yann Mathet, Antoine Widl{\"o}cher, and Jean-Philippe M{\'e}tivier. 2015.
\newblock The unified and holistic method gamma ($\gamma$) for inter-annotator
  agreement measure and alignment.
\newblock \emph{Computational Linguistics}, 41(3):437--479.

\bibitem[{Meyer et~al.(2014)Meyer, Mieskes, Stab, and
  Gurevych}]{meyer-EtAl:2014:ColingDemo}
Christian~M. Meyer, Margot Mieskes, Christian Stab, and Iryna Gurevych. 2014.
\newblock {DKPro} agreement: An open-source {J}ava library for measuring
  inter-rater agreement.
\newblock In \emph{Proceedings of the International Conference on Computational
  Linguistics}, COLING~'14, pages 105--109, Dublin, Ireland.

\bibitem[{Miller(1939)}]{Miller}
Clyde~R. Miller. 1939.
\newblock {The Techniques of Propaganda}.
\newblock From ``How to Detect and Analyze Propaganda,'' an address given at
  Town Hall.
\newblock The Center for learning.

\bibitem[{Nadeau and Sekine(2007)}]{Nadeau:07}
David Nadeau and Satoshi Sekine. 2007.
\newblock A survey of named entity recognition and classification.
\newblock \emph{Lingvisticae Investigationes}, 30(1):3--26.

\bibitem[{Potthast et~al.(2010)Potthast, Stein, Barr{\'o}n-Cede{\~n}o, and
  Rosso}]{Potthast2010a}
Martin Potthast, Benno Stein, Alberto Barr{\'o}n-Cede{\~n}o, and Paolo Rosso.
  2010.
\newblock An evaluation framework for plagiarism detection.
\newblock In \emph{Proceedings of the International Conference on Computational
  Linguistics}, COLING~'10, pages 997--1005, Beijing, China.

\bibitem[{Rashkin et~al.(2017)Rashkin, Choi, Jang, Volkova, and
  Choi}]{rashkin-EtAl:2017:EMNLP2017}
Hannah Rashkin, Eunsol Choi, Jin~Yea Jang, Svitlana Volkova, and Yejin Choi.
  2017.
\newblock Truth of varying shades: Analyzing language in fake news and
  political fact-checking.
\newblock In \emph{Proceedings of the 2017 Conference on Empirical Methods in
  Natural Language Processing}, EMNLP~'17, pages 2931--2937, Copenhagen,
  Denmark.

\bibitem[{Richter(2017)}]{Richter2017}
Monika~L Richter. 2017.
\newblock The {K}remlin's platform for `useful idiots' in the {W}est: An
  overview of {RT}'s editorial strategy and evidence of impact.
\newblock Technical report, Kremlin Watch.

\bibitem[{Suprabandari(2007)}]{Suprabandari2007}
Francisca Niken~Vitri Suprabandari. 2007.
\newblock {American propaganda in John Steinbeck's The Moon is Down}.
\newblock Master's thesis, Sanata Dharma University, Yogyakarta, Indonesia.

\bibitem[{Teninbaum(2009)}]{Aper2009}
Gabriel~H Teninbaum. 2009.
\newblock Reductio ad {H}itlerum: Trumping the judicial {N}azi card.
\newblock \emph{Michigan State Law Review}, page 541.

\bibitem[{Tjong Kim~Sang(2002)}]{TjongKimSang:2002:ICS:1118853.1118877}
Erik~F. Tjong Kim~Sang. 2002.
\newblock Introduction to the {CoNLL}-2002 shared task: Language-independent
  named entity recognition.
\newblock In \emph{Proceedings of the 6th Conference on Natural Language
  Learning}, CoNLL~'02, pages 155--158, Taipei, Taiwan.

\bibitem[{Tjong Kim~Sang and
  De~Meulder(2003)}]{TjongKimSang:2003:ICS:1119176.1119195}
Erik~F. Tjong Kim~Sang and Fien De~Meulder. 2003.
\newblock Introduction to the {CoNLL}-2003 shared task: Language-independent
  named entity recognition.
\newblock In \emph{Proceedings of the 7th Conference on Natural Language
  Learning}, CoNLL '03, pages 142--147, Edmonton, Canada.

\bibitem[{Torok(2015)}]{Torok2015}
Robyn Torok. 2015.
\newblock {Symbiotic radicalisation strategies: Propaganda tools and neuro
  linguistic programming}.
\newblock In \emph{Proceedings of the Australian Security and Intelligence
  Conference}, pages 58--65, Perth, Australia.

\bibitem[{Tsai et~al.(2006)Tsai, Wu, Chou, Lin, He, Hsiang, Sung, and
  Hsu}]{Tsai:06}
Richard Tzong-Han Tsai, Shih-Hung Wu, Wen-Chi Chou, Yu-Chun Lin, Ding He, Jieh
  Hsiang, Ting-Yi Sung, and Wen-Lian Hsu. 2006.
\newblock {Various criteria in the evaluation of biomedical named entity
  recognition.}
\newblock \emph{BMC bioinformatics}, 7:92.

\bibitem[{Walton(1996)}]{Walton1996}
Douglas Walton. 1996.
\newblock \emph{The straw man fallacy}.
\newblock Royal Netherlands Academy of Arts and Sciences.

\bibitem[{Weston(2018)}]{Weston2000}
Anthony Weston. 2018.
\newblock \emph{A rulebook for arguments}.
\newblock Hackett Publishing.

\end{thebibliography}
\bibliographystyle{acl_natbib}

\appendix
\counterwithin{figure}{section}
\section{Annotation Guidelines} 
\label{app:annotation}

\begin{figure*}
\centering
\fbox{%
\begin{minipage}[t]{.48\textwidth}
\footnotesize
We aim at identifying propagandistic techniques in news articles.
We provide you with a news article and a flowchart to guide you through the identification of propaganda techniques.
The definition of each technique is shown when setting the mouse pointer on the name of the technique in the flowchart.
You are free to annotate single words, phrases, or sentences, but we encourage you to select the minimal amount of text in which the propaganda technique appears. 
\vspace{-5mm}
\begin{enumerate}\setlength\itemsep{0em}
\item Let us look at the flowchart [below]
	\item Let us look at an example which includes four propaganda techniques [cf.\ Figure~1]
\end{enumerate}
 
\begin{itemize}	\setlength\itemsep{0em}
	\item {\bf Name calling}: the democrats are being called "babies"
	\item {\bf Black-and-white fallacy}: obstruction vs progress
	\item {\bf Loaded language}: stupid, petty, killing
\end{itemize}
\end{minipage}
\hspace{0.02\textwidth}
\begin{minipage}[t]{.48\textwidth}
 \footnotesize
 \begin{itemize}	\setlength\itemsep{0em}
 \item {\bf Exaggeration}: killing a grandma, stomaching the presence of a person
\end{itemize}
\vspace{-3mm}
Use the flowchart as your guide to spot propaganda.
If you are not sure about a propaganda technique (any rounded box in the flowchart), just click on it and a new page will open with explanations and examples when necessary. [cf.\ Section~2] 

{\bf TIPS}
\vspace{-3mm}
\begin{itemize}	\setlength\itemsep{0em}
	\item Some sentences might be tricky. Please try to select the right technique(s)
	\item Your emotions have nothing to do with the articles, as you are requested to spot propagandistic techniques, not their message: try to distance yourself from the contents and avoid being biased.
	\item One text fragment may include more than one technique at the same time
\end{itemize}

\end{minipage}
}
\includegraphics[scale=0.47]{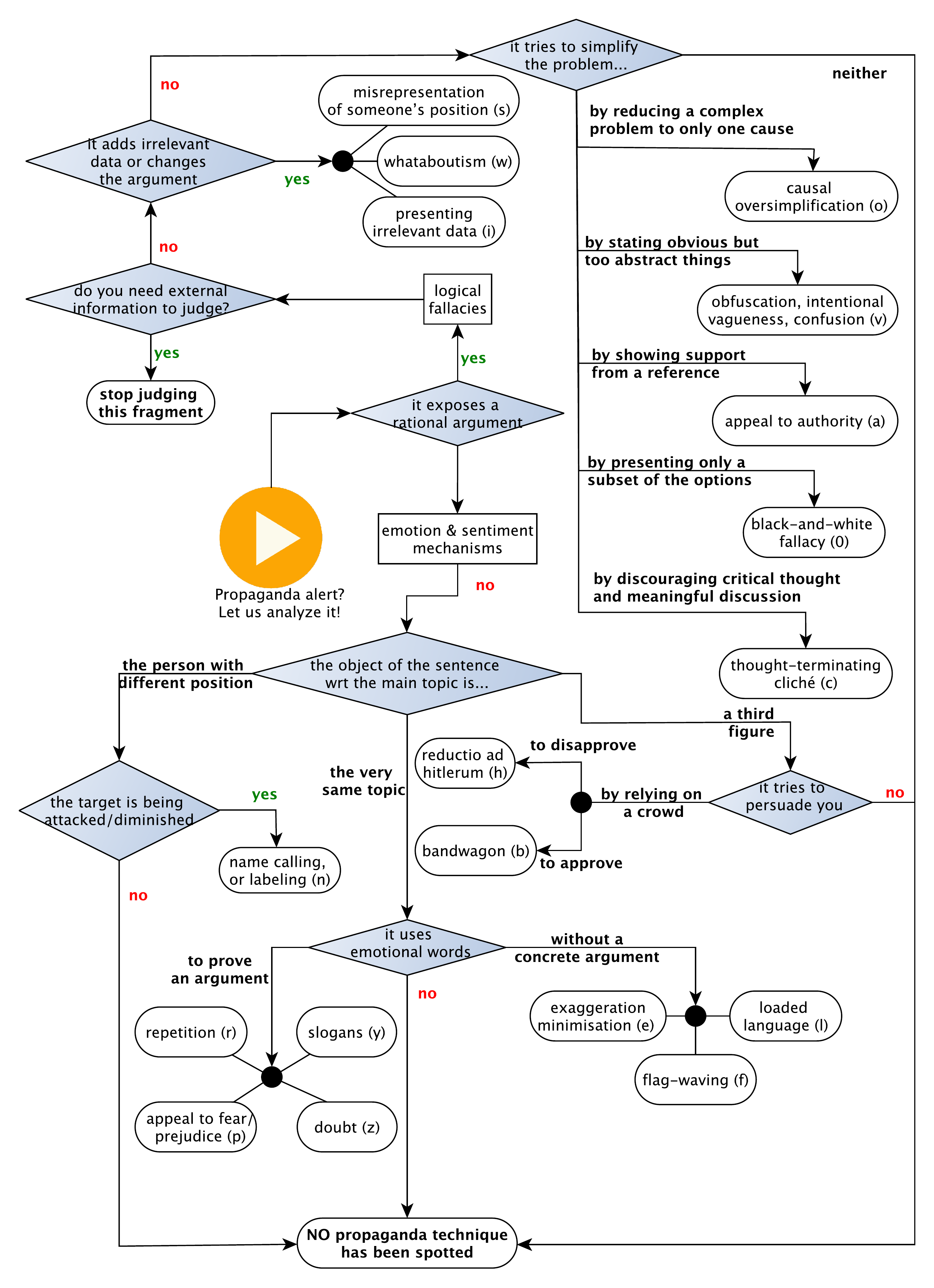}
 \caption{Instructions as provided to the professional annotators in the Anafora-based annotation process (top). Flowchart to drive the identification of propaganda techniques (bottom). 
 \label{fig:flowchart}}
\end{figure*}

We provided the definitions in Section~2 together with some examples and an annotation schema, to our professional annotators, so that they could manually annotate news articles. The annotators performed their task following the instructions displayed in Figure~\ref{fig:flowchart}. In order to help them, we built the flowchart displayed in the same figure. It partitions the set of techniques hierarchically and can be traversed by answering a series of questions. 
These instructions and the flowchart were always available to the annotators, next to the annotation interface (cf.\ Figure~1). 
As an example of the process for generating the questions, the first subdivision is inspired by the following description of propaganda: ``Propaganda comes in many forms, but you can recognize it by [\ldots] the use of faulty reasoning and/or emotional appeals''. The description distinguishes between logical fallacies and techniques appealing to emotions.

\end{document}